\documentclass[journal]{IEEEtran}
\usepackage{times}
\usepackage{soul}
\usepackage{url}
\usepackage[hidelinks]{hyperref}
\usepackage[utf8]{inputenc}
\usepackage[small]{caption}
\usepackage{graphicx}
\usepackage{amsmath}
\usepackage{amsthm}
\usepackage{amssymb}
\usepackage{booktabs}
\usepackage[noend]{algpseudocode}
\usepackage{algorithm}
\usepackage{algorithmicx}
\usepackage{multirow}
\usepackage{color,xcolor}
\usepackage{comment}

\usepackage{multirow}
\usepackage{bbding}

\ifCLASSINFOpdf
\else
\fi
\begin{document}
\title{HiREN: Towards Higher Supervision Quality for
Better Scene Text Image Super-Resolution}

% \author{Michael~Shell,~\IEEEmembership{Member,~IEEE,}
%         John~Doe,~\IEEEmembership{Fellow,~OSA,}
%         and~Jane~Doe,~\IEEEmembership{Life~Fellow,~IEEE}% <-this % stops a space
% \thanks{M. Shell was with the Department
% of Electrical and Computer Engineering, Georgia Institute of Technology, Atlanta,
% GA, 30332 USA e-mail: (see http://www.michaelshell.org/contact.html).}% <-this % stops a space
% \thanks{J. Doe and J. Doe are with Anonymous University.}% <-this % stops a space
% \thanks{Manuscript received April 19, 2005; revised August 26, 2015.}
\author{Minyi Zhao, Yi Xu, Bingjia Li, Jie Wang, Jihong Guan, and Shuigeng Zhou,~\IEEEmembership{Senor Member,~IEEE}
% \thanks{\IEEEauthorrefmark{1}Corresponding author.}
\thanks{Manuscript received July 25, 2023.}
\thanks{Minyi Zhao, Yi Xu, Bingjia Li and Shuigeng Zhou (Corresponding author) are with Shanghai Key Lab of Intelligent Information Processing, and School of Computer Science, Jiangwan Campus, Fudan University, 2005 Songhu Road, Shanghai, 200438, China. E-mail: zhaomy20@fudan.edu.cn; yxu17@fudan.edu.cn; bjli20@fudan.edu.cn; sgzhou@fudan.edu.cn}
\thanks{Jie Wang is with ByteDance Inc, Beijing 100098, China. E-mail: wangjie.bernard@bytedance.com}
\thanks{Jihong Guan is with Department of Computer Science and Technology, Tongji University, 4800 Caoan Road, Shanghai, 201804, China. E-mail: jhguan@tongji.edu.cn
}
}


\markboth{Journal of \LaTeX\ Class Files,~Vol.~14, No.~8, August~2015}%
{Shell \MakeLowercase{\textit{et al.}}: Bare Demo of IEEEtran.cls for IEEE Journals}



\maketitle

\begin{abstract}
Scene text image super-resolution (STISR) is an important pre-processing technique for text recognition from low-resolution scene images. Nowadays, various methods have been proposed to extract text-specific information from high-resolution (HR) images to supervise STISR model training. However, due to uncontrollable factors  (\textit{e.g.} shooting equipment, focus, and environment) in manually photographing HR images, the quality of HR images cannot be guaranteed, which unavoidably impacts STISR performance. Observing the quality issue of HR images, in this paper we propose a novel idea to boost STISR by first enhancing the quality of HR images and then using the enhanced HR images as supervision to do STISR. Concretely, we develop a new STISR framework, called \textbf{H}igh-\textbf{R}esolution \textbf{EN}hancement (HiREN) that consists of two branches and a quality estimation module. The first branch is developed to recover the low-resolution (LR) images, and the other is an \emph{HR quality enhancement} branch aiming at generating high-quality (HQ) text images based on the HR images to provide more accurate supervision to the LR images. As the degradation from HQ to HR may be diverse, and there is no pixel-level supervision for HQ image generation, we design a kernel-guided enhancement network to handle various degradation, and exploit the feedback from a recognizer and text-level annotations as weak supervision signal to train the HR enhancement branch. Then, a \emph{quality estimation module} is employed to evaluate the qualities of HQ images, which are used to suppress the erroneous supervision information by weighting the loss of each image. Extensive experiments on TextZoom show that HiREN can work well with most existing STISR methods and significantly boost their performances.
% Code is available in the supplementary material.
\end{abstract}


% \fi
\IEEEpeerreviewmaketitle

\begin{figure*}
	\begin{center}
		\includegraphics[width=0.85\linewidth]{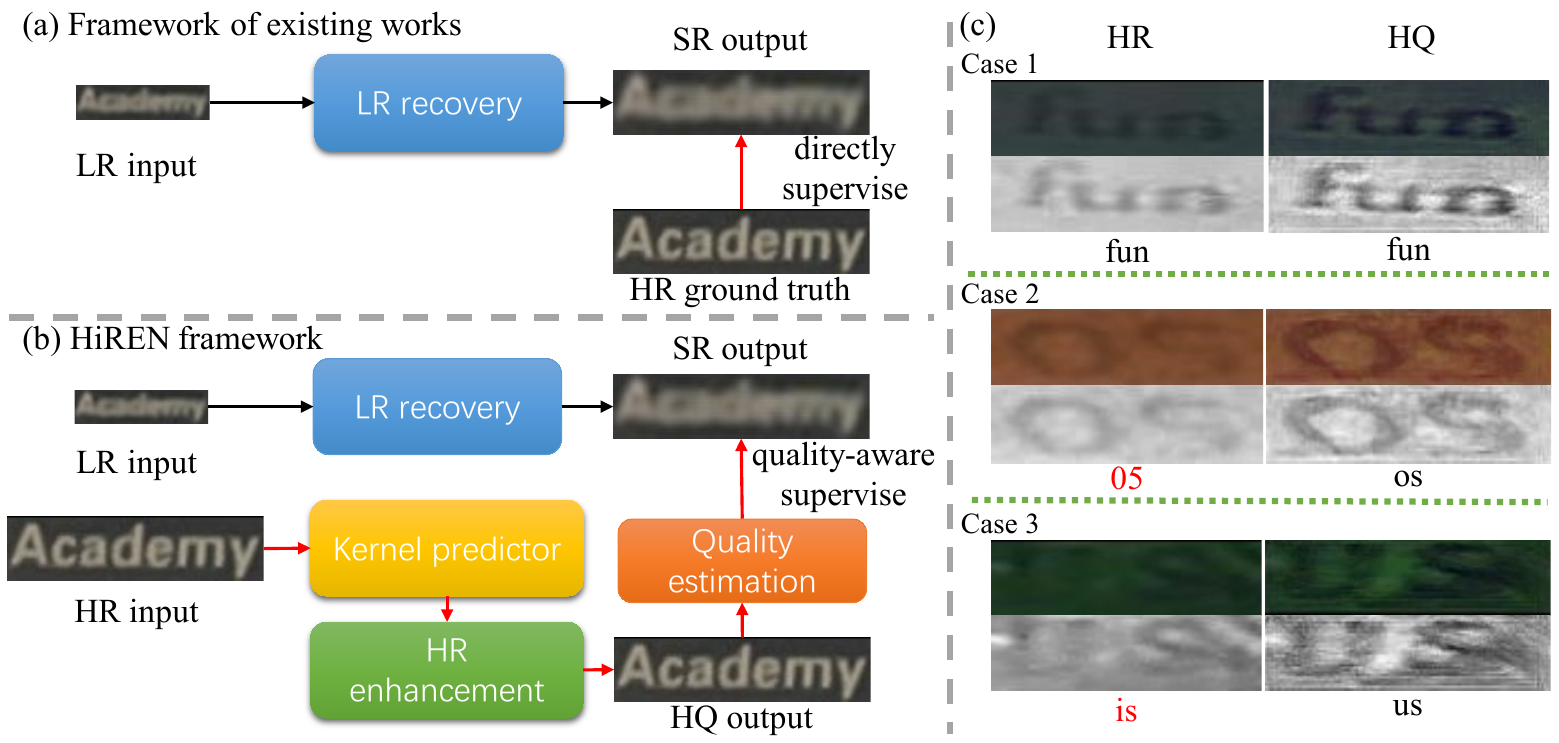}
	\end{center}
	\caption{Overview of existing STISR approaches and our method, and examples illustrating the quality problem of HR images. (a) The framework of existing STISR methods; (b) The HiREN framework; (c) Some examples of low-quality HR images and their enhanced results (HQ) by our method, as well as the recognized results. For each case, the 1st row shows HR and HQ images, the 2nd row presents the normalized HR and HQ images to highlight their visual differences, and the 3rd row gives the recognized characters: red indicates incorrectly recognized, and black means correctly recognized.}
	\label{fig-intro}
\end{figure*}

\section{Introduction}
Scene text recognition~\cite{chen2021benchmarking,chen2021survey} (STR), which aims at recognizing texts from scene images has wide applications in scene text based image understanding (\textit{e.g.} auto-driving~\cite{zhang2020street}, TextVQA~\cite{singh2019towards}, Doc-VQA~\cite{mathew2021docvqa}, and ViteVQA~\cite{zhaotowards}). Despite the fact that STR has made great progress with the rapid blossom of deep learning in recent years, performance of text recognition from low-resolution (LR) text images is still unsatisfactory~\cite{wang2020scene}. Therefore, scene text image super-resolution (STISR)~\cite{chen2021scene,chen2022text,wang2020scene} is gaining popularity as a pre-processing technique to recover the missing details in LR images for boosting text recognition performance as well as the visual quality of the scene texts.

As shown in Fig.~\ref{fig-intro}(a), recent STISR works usually try to directly capture pixel-level (via $L1$ or $L2$ loss) or text-specific information from high-resolution (HR) text images to supervise the training of STISR models. For instance, Gradient profile loss~\cite{wang2020scene} calculates the gradient fields of HR images as ground truth for sharpening the boundaries of the super-resolution (SR) images. PCAN~\cite{zhao2021scene} is proposed to learn sequence-dependent features and high-frequency information of the HR images to better reconstruct SR text images. STT~\cite{chen2021scene} exploits character-level attention maps from HR images to assist the recovery. \cite{nakaune2021skeleton} and TG~\cite{chen2022text} extract stroke-level information from HR images through specific networks to provide  more fine-grained supervision information. \cite{ma2023text,ma2022text,zhao2022c3} additionally introduce external modules to extract various text-specific clues to facilitate the recovery and use the supervision from HR images to finetune their modules.

Although various techniques that extract information from the HR images have been proposed to improve the recognition accuracy, they all assume that the HR images are completely trustworthy, which is actually not true, due to the uncontrollable factors (e.g. shooting equipment, focus, and environment) in manually photographing the HR images. As shown in Fig.~\ref{fig-intro}(c), the HR images may suffer from blurring (the 1st and 2nd cases) and low-contrast (the 3rd case), which unavoidably impacts the performance of STISR. In the worst case, these quality issues may cause the failure of recognition on HR images and lead to wrong supervision information. What is worse, the HR quality problem in real world is absolutely not negligible, as the recognition accuracy on HR images can be as low as 72.4\% (see Tab.~\ref{tab:final result}).

Considering the fact that improving the photographing of LR/HR images and eliminating environmental impacts are extremely expensive (if not impossible) in the wild, and applying huge models for extracting more accurate information is also time-consuming and costly, in this paper we propose a novel solution to advance STISR by first enhancing the quality of HR images and then using the enhanced HR images as supervision to perform STISR. To this end, we develop a new, general and easy-to-use STISR framework called \textbf{H}igh-\textbf{R}esolution \textbf{EN}chancement (HiREN) to improve STISR by providing more accurate supervision.
In particular, as shown in Fig.~\ref{fig-intro}(b), besides the typical LR recovery branch, HiREN additionally introduces an HR enhancement branch that aims at improving the quality of HR images and a quality estimation (QE) module to conduct a quality-aware supervision. Here, the resulting high-quality (HQ) images, instead of the HR images as in existing works, are used to supervise the LR recovery branch. Note that the degradation from HQ to HR is unknown, and there is no explicit supervision for HR enhancement, existing STISR approaches are not able to solve the task of HR enhancement. To tackle these problems, on the one hand, we introduce a degradation kernel predictor to generate the degradation kernel and then use this kernel as a clue to enhance various degraded HR images. On the other hand, we exploit the feedback of a scene text recognizer and text-level annotations as weak supervision signal to train the HR enhancement branch. What is more, to suppress the erroneous supervision information, a quality estimation (QE) module is proposed to evaluate the quality of the HQ images through the normalized Levenshtein similarity~\cite{levenshtein1966binary} of the recognized text and the ground truth, and then use this quality estimation to weight the loss of each HQ image.

Such design above offers our method four-fold advantages:
\begin{itemize}
  \item \emph{General}. Our framework can work with most existing STISR approaches in a plug-and-play manner.
  \item \emph{Easy-to-use}. After training the HR enhancement branch, our method can be plugged online to the training of existing techniques easily.
  \item \emph{Efficient}. HiREN does not introduce additional cost during inference. What is more, HiREN can also be deployed offline by caching all the enhanced HR images. This offline deployment does not introduce any additional training cost.
  \item \emph{High-performance}. Our method can significantly boost the performances of existing methods.
\end{itemize}

Contributions of this paper are summarized as follows:
\begin{itemize}
    \item We propose a novel approach for STISR. To the best of our knowledge, this is the first work to consider and exploit the quality of HR images in STISR. That is, different from existing approaches that extract various text-specific information, Our work pioneers the exploration of the quality issue of HR images.
    \item We develop a general, efficient and easy-to-use \textbf{H}igh-\textbf{R}esolution \textbf{EN}hancement (HiREN) framework to boost STISR by improving the supervision information from the HR images.
    \item We conduct extensive experiments on TextZoom, which show that HiREN is compatible with most existing STISR methods and can significantly lift their performances.
\end{itemize}

The rest of this paper is organized as follows: Section~\ref{sec:related-work} surveys related works and highlights the differences between our method and the existing ones; Section~\ref{sec:method} presents our method in detail; Section~\ref{sec:performance-evaluation} introduce the experimental results of our method and performance comparisons with existing methods; Section~\ref{sec:discussion} further discusses the quality issues of HR images, error cases and limitations of the proposed method; Section~\ref{sec:conclusion} concludes the paper while pinpointing some issues of future study.

\section{Related Work}\label{sec:related-work}
In this section, we briefly review the super-resolution techniques and some typical scene text recognizers.
According to whether exploiting text-specific information from HR images, recent STISR methods can be roughly divided into two groups: generic super-resolution approaches and scene text image super-resolution approaches.

\subsection{Generic Image Super-Resolution}
Generic image super-resolution methods~\cite{yang2019deep,tian2020coarse,li2022real,jiang2022single} usually recover LR images through pixel information from HR images captured by pixel loss functions. In particular, SRCNN~\cite{dong2015image} is a three-layer convolutional neural network. \cite{xu2017learning} and SRResNet~\cite{ledig2017photo} adopt generative adversarial networks to generate distinguishable images. \cite{pandey2018binary} employs convolutional layers, transposed convolution and sub-pixel convolution layers to extract and upscale features. RCAN~\cite{zhang2018image} and SAN~\cite{dai2019second} introduce attention mechanisms to boost the recovery. Nowadays, transformer-structured approaches~\cite{li2021efficient,liang2021swinir,wang2022uformer} are proposed to further advance the task of generic image super-resolution. Nevertheless, these approaches ignore text-specific properties of the scene text images, which leads to low recognition performance when applied to STISR.

\subsection{Scene Text Image Super-Resolution}
Recent approaches focus on extracting various text-specific information from the HR images, which is then utilized to supervise model training. Specifically, \cite{fang2021tsrgan,wang2019textsr} calculate text-specific losses to boost performance. \cite{mou2020plugnet} proposes a multi-task framework that jointly optimizes recognition and super-resolution branches. \cite{wang2020scene} introduces TSRN and gradient profile loss to capture sequential information of text images and gradient
fields of HR images for sharpening the texts. PCAN~\cite{zhao2021scene} is proposed to learn sequence-dependent and high-frequency information of the reconstruction. STT~\cite{chen2021scene} makes use of character-level information from HR images extracted by a pre-trained transformer recognizer to conduct a text-focused super-resolution. \cite{qin2022scene} proposes a content perceptual loss to extract multi-scale
text recognition features to conduct a content aware supervision. TPGSR~\cite{ma2023text}, TATT~\cite{ma2022text}, and C3-STISR~\cite{zhao2022c3} extract text-specific clues to guide the super-resolution. In particular, TPGSR is the first method that additionally introduces a scene text recognizer to provide text priors. Then, the extracted priors are fed into the super-resolution to iteratively benefit the super-resolution. TATT~\cite{ma2022text} introduces a transformer-based module, which leverages global attention mechanism, to exert the semantic guidance of text prior to the text reconstruction process. C3-STISR~\cite{zhao2022c3} is proposed to learn triple clues, including recognition clue from a STR, linguistical clue from a language model, and a visual clue from a skeleton painter to rich the representation of the text-specific clue. TG~\cite{chen2022text} and \cite{nakaune2021skeleton} exploit stroke-level information from HR images via stroke-focused module and skeleton loss for more fine-grained super-resolution. Compared with generic image super-resolution approaches, these methods greatly advance the recognition accuracy through various text-specific information extraction techniques. Nevertheless, they all assume that HR images are completely trustable, which is actually not true in practice. As a result, their extracted supervision information may be erroneous, which impacts the STISR performance. Since HiREN applies these methods to implement the LR recovery branch,  to elaborate the differences among various super-resolution techniques in this paper, we give a summary of these methods in Tab.~\ref{tab:lrrecovery} on three major aspects: how their super-resolution blocks and loss functions are designed, and whether they use iterative super-resolution technique to boost the performance.
\begin{table}[t]
%\begin{subtable}
\caption{Differences between typical STISR methods from three aspects: super-resolution block, loss function, and whether this method is iterative or not.}
\centering
\scalebox{0.92}{
\begin{tabular}{c|ccc}
\hline
Method & Super-resolution block & Loss function $\mathcal{L}_{LR}$& Iterative\\
 \hline
SRCNN~\cite{dong2015image} & SRCNN~\cite{dong2015image} & MSE & $\times$\\
SRResNet~\cite{ledig2017photo} & SRResNet~\cite{ledig2017photo} & MSE & $\times$\\
TSRN~\cite{wang2020scene} & SRB~\cite{wang2020scene} & Gradient profile loss~\cite{wang2020scene} & $\times$\\
PCAN~\cite{zhao2021scene} & PCA~\cite{zhao2021scene} & Edge guidance loss~\cite{zhao2021scene} & $\times$\\
STT~\cite{chen2021scene} & TBSRN~\cite{chen2021scene} & Text-focused loss~\cite{chen2021scene} & $\times$\\
TPGSR~\cite{ma2023text} & SRB~\cite{wang2020scene} & Gradient profile loss~\cite{wang2020scene} & $\checkmark$ \\
TG~\cite{chen2022text} & SRB~\cite{wang2020scene} & Stroke-focused loss~\cite{chen2022text} & $\times$\\
\hline
\end{tabular}}
\label{tab:lrrecovery}
\end{table}
\subsection{Scene Text Recognition}
Scene text recognition (STR)~\cite{bai2018edit,chen2021benchmarking,chen2021survey,cheng2017focusing,cheng2018aon} has made great progress in recent years. Specifically, CRNN~\cite{shi2016end} takes CNN and RNN as the encoder and employs a CTC-based~\cite{graves2006connectionist} decoder to maximize the probabilities of paths that can reach the ground truth. ASTER~\cite{shi2018aster} introduces a spatial transformer network (STN)~\cite{jaderberg2015spatial} to rectify irregular text images. MORAN~\cite{luo2019moran} proposes a multi-object rectification network. \cite{hu2020gtc,sheng2019nrtr,yu2020towards} propose novel attention mechanisms. AutoSTR~\cite{zhang2020efficient} searches backbone via neural architecture search (NAS)~\cite{elsken2019neural}. More recently, semantic-aware~\cite{qiao2020seed,yu2020towards}, transformer-based~\cite{atienza2021vision}, linguistics-aware~\cite{fang2021read,wang2021two}, and efficient~\cite{du2022svtr,bautista2022parseq} approaches are proposed to further boost the performance. Although these methods are able to handle irregular, occluded, and incomplete text images, they still have difficulty in recognizing low-resolution images. For example, as can be seen in Sec.~\ref{sec:exp:sota}, CRNN, MORAN, and ASTER only achieve the recognition accuracy of 27.3\%, 41.1\% and 47.2\% respectively when directly using LR images as input. What is more, finetuning these recognizers is insufficient to accurately recognize texts from LR images, as reported in \cite{wang2020scene}. Therefore, a pre-processor is required for recovering the details of low-resolution images.

\subsection{Difference between Our Method and Existing STISR Works}
The motivation of HiREN is totally different from that of existing STISR approaches. As described above, existing methods focus on extracting text-specific information from HR images to supervise STISR. On the contrary, HiREN first lifts the quality of HR images, then uses the enhanced images to supervise STISR. This allows HiREN to work with most existing STISR approaches and boost their recognition performances in a general, economic and easy-to-use way.

\section{Method}\label{sec:method}
Here, we first give an overview of our framework HiREN, then briefly introduce the LR recovery branch. Subsequently, we present the HR enhancement branch and the quality estimation module in detail, followed by the usage of HiREN.
\begin{figure*}[t]
	\begin{center}
		\includegraphics[width=1.0\linewidth]{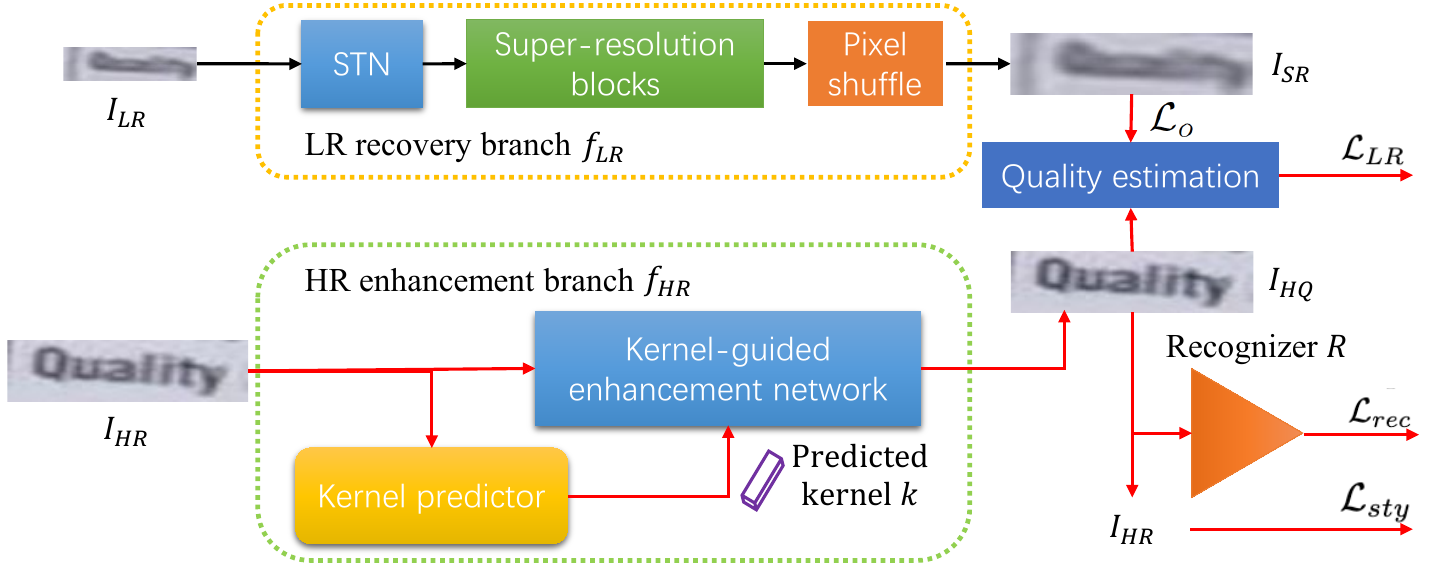}
	\end{center}
	\caption{The framework of HiREN. Red lines are valid only during training.}
	\label{fig-pipeline}
\end{figure*}

\subsection{Overview}
Given a low-resolution (LR) image $I_{LR} \in \mathbb{R}^{C\times N}$ . Here, $C$ is the number of channels of the image, $N = H \times W$ is the collapsed spatial dimension, $H$ and $W$ are the height and width of image $I_{LR}$. Our aim is to produce a super-resolution (SR) image $I_{SR} \in \mathbb{R}^{C \times (4 \times N)}$ with the magnification factor of $\times 2$. Fig.~\ref{fig-pipeline} shows the architecture of our framework HiREN, which is composed of two major branches: the \emph{LR recovery branch} $f_{LR}$ that takes $I_{LR}$ as input to generate a super-resolution image $I_{SR}=f_{LR}(I_{LR})$ and a corresponding loss $\mathcal{L}_{o}$, and the \emph{HR enhancement branch} $f_{HR}$ that takes $I_{HR}$ as input to generate a high-quality (HQ) image $I_{HQ}=f_{HR}(I_{HR})$ where $I_{HQ} \in \mathbb{R}^{C \times (4 \times N)}$, and a \emph{quality estimation module} $f_{QE}$ that takes $I_{HQ}$ and $\mathcal{L}_o$ as input to compute a quality-aware loss $\mathcal{L}_{LR}$ to supervise the LR branch:
\begin{equation}
\label{eq-overview}
    \mathcal{L}_{LR} = f_{QE}(I_{HQ},\mathcal{L}_o).
\end{equation}
During inference, $f_{HR}$ and $f_{QE}$ are removed. Thus, HiREN does not introduce extra inference cost.

\subsection{LR Recovery Branch}
In HiREN, the LR recovery branch can be one of the existing STISR approaches. As shown in Fig.~\ref{fig-pipeline}, these methods usually work in the following way: 1) Start with a spatial transformer network (STN)~\cite{jaderberg2015spatial} since in the TextZoom dataset~\cite{wang2020scene} the HR-LR pairs are manually cropped and matched by humans, which may incur several pixel-level offsets. 2) Several super-resolution blocks are used to learn sequence-dependent information of text images. 3) A pixel shuffle module is employed to reshape the super-resolved image. 4) Various loss functions are served as $\mathcal{L}_{o}$ to extract text-specific information from ground truth ($I_{HR}$ in existing works, $I_{HQ}$ in HiREN) to provide the supervision. To elaborate the differences among various LR branches tested in this paper, we give a summary of these methods in Tab.~\ref{tab:lrrecovery}.

% As aforementioned, existing works mainly focus on designing super-resolution blocks (\textit{i.e.,} SRB in \cite{wang2020scene}, TBSRN in \cite{chen2021scene}) or extracting text-specific supervision information (\textit{i.e.,} text-focused loss in \cite{chen2021scene} and stroke-focused loss in \cite{chen2022text}). In Tab.~\ref{tab:lrrecovery}, we summarize some typical existing methods to better demonstrate the LR recovery branch.
As the motivation of HiREN is totally different from that of the existing methods, our method can work with most of them and significantly improve their performances.

\subsection{HR Enhancement Branch}
\subsubsection{Overall introduction.}
The enhancement of HR images is a challenging task, where the challenges lie in two aspects that will be detailed in the sequel. Formally, the HR image $I_{HR}$ and the corresponding HQ image $I_{HQ}$ we are pursuing are connected by a degradation model as follows:
\begin{equation}
\label{eq:degra}
    I_{HR}= k \otimes I_{HQ} + n,
\end{equation}
where $\otimes$ denotes the convolution operation, $k$ is the degradation kernel, and $n$ is the additive noise that follows Gaussian distribution in real world applications~\cite{gu2019blind,zhang2018learning}.
Different from the degradation from $I_{HR}$ to $I_{LR}$ where the kernel is determined by lens zooming, unfortunately, the degradation $k$ of $I_{HQ}$ is unknown. As shown in Fig.~\ref{fig-intro}(c), such degradation can be but not limited to blurring (the 1st and 2nd cases) and low-contrast (the 3rd case). What is more, we also lack pixel-level supervision information of $I_{HQ}$. These two challenges make existing STISR methods unable to enhance $I_{HR}$. To cope with the first challenge, here we adopt blind image deblurring techniques~\cite{michaeli2013nonparametric,yuan2018unsupervised,zhang2018learning,gu2019blind} to boost the recovery of $I_{HR}$.
Specifically, as shown in Fig.~\ref{fig-pipeline}, our HR enhancement branch consists of two components: a \emph{kernel predictor} $P$ and a \emph{kernel-guided enhancement network} $f_{ke}$. The kernel predictor aims at estimating the degradation kernel $k$ (\textit{i.e.,} $k = P(I_{HR})$ where $k\in \mathbb{R}^{d}$, and $d$ is the size of the kernel), while the kernel-guided enhancement network takes the predicted kernel and $I_{HR}$ as input to conduct a kernel-guided enhancement: $I_{HQ}=f_{ke}(I_{HR},k)$. The predicted kernel is utilized as a clue to strengthen the model's ability to handle various degradation and boost the recovery of HR images. As for the second challenge, we introduce a pre-trained scene text recognizer $R$ to provide the supervision for generating more recognizable HQ images. And after training the HR enhancement branch $f_{HR}$, HiREN uses the trained $f_{HR}$ to generate HQ images, which are exploited for training the LR recovery branch.

\begin{figure}[t]
	\begin{center}
		\includegraphics[width=1.0\linewidth]{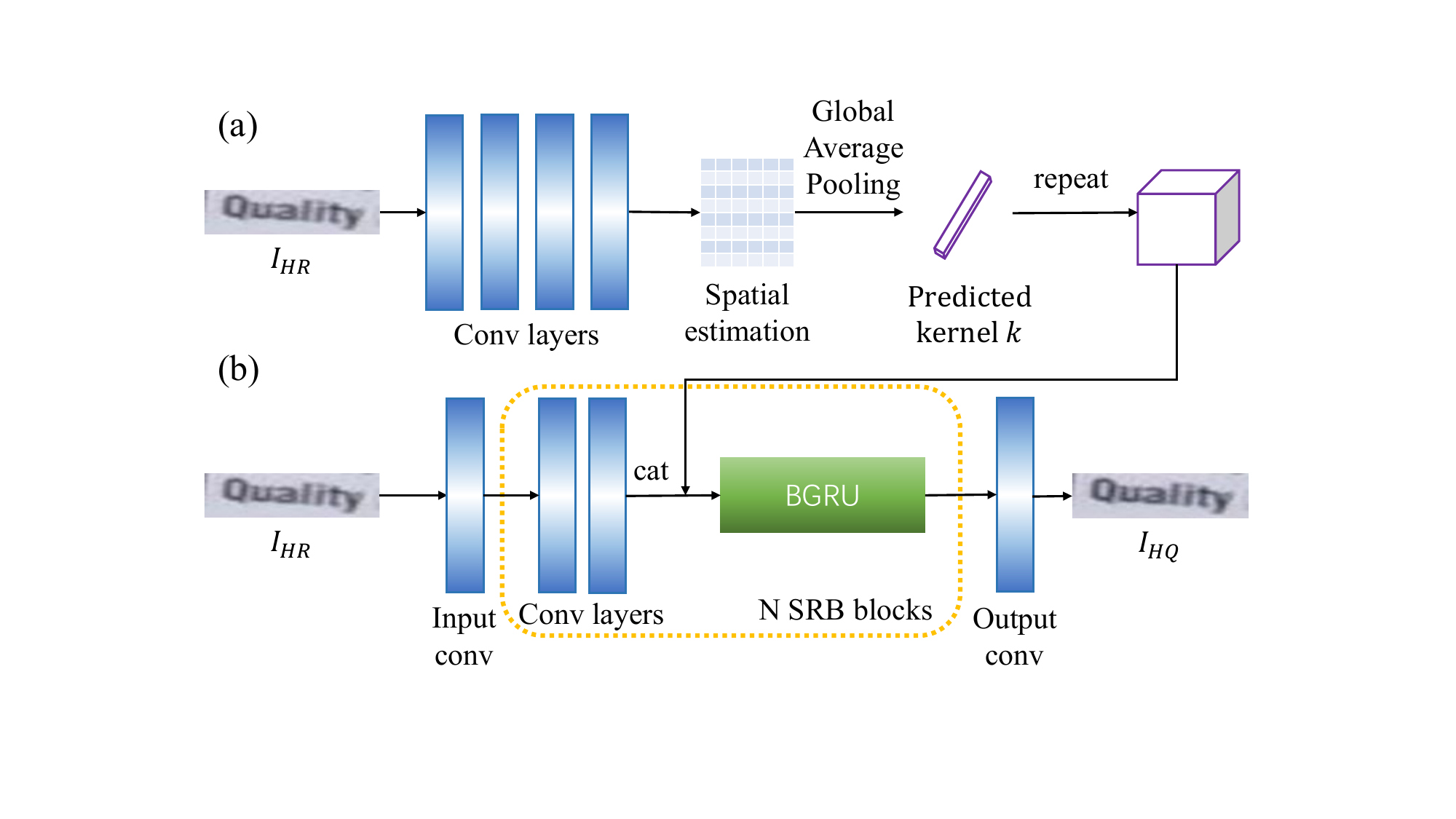}
	\end{center}
	\caption{The structure of the HR enhancement branch, which consists of two components: (a) the kernel predictor $P$, and (b) the kernel-guided enhancement network $f_{ke}$.}
	\label{fig-detail}
\end{figure}

\subsubsection{The kernel predictor.}
As shown in Fig.~\ref{fig-detail}, to generate a prediction of the degradation kernel, we first utilize convolution layers to obtain a spatial estimation of the kernel. Then, we employ global average pooling~\cite{lin2013network} to output the global prediction by evaluating the spatial mean value. Thus, we can get the prediction of the kernel of size $\mathbb{R}^d$, in a simple yet effective way.

\subsubsection{The kernel-guided enhancement network.}
As shown in Fig.~\ref{fig-detail}, our kernel-guided enhancement network is designed in the following way: 1) Start with an input convolution to change the channel number from $C$ to $C'$. 2) Repeat $N$ modified SRB blocks~\cite{wang2020scene}. Each block consists of two convolution layers and one Bi-directional GRU~\cite{chung2014empirical} (BGRU) to handle sequential text images. At this step, we first stretch the predicted kernel $k$ to pixel shape, then concatenate
the pixel kernel with the feature map extracted by convolution layers at channel dimension. 3) An output convolution is applied to getting the final enhanced HQ image $I_{HQ}$.

\subsubsection{Loss functions.}
\label{sec:loss:hr_branch}
Here, we design the loss functions of the HR enhancement branch $f_{HR}$. As shown in Fig.~\ref{fig-pipeline}, there are two loss functions in $f_{HR}$. The first one is the recognition loss $\mathcal{L}_{rec}$ that is used to make the enhanced image $I_{HQ}$ to be more easily recognized than $I_{HR}$. It is provided by a pre-trained recognizer $R$ and the text-level annotation of $I_{HR}$. Suppose the encoded text-level annotation is $p_{GT} \in \mathbb{R}^{L\times |\mathcal{A}|}$, where $L$ is the max prediction length of recognizer $R$, and $|\mathcal{A}|$ denotes the length of the alphabet $\mathcal{A}$. Then, the recognition loss can be evaluated by
\begin{equation}
    \label{eq:rec}
    \mathcal{L}_{rec} = - \sum_{j=0}^{L}p_{GT}^j log(R(I_{HQ})^j),
\end{equation}
which is the cross entropy of $p_{GT}$ and $R(I_{HQ})$. Beside the recognition loss, it is essential to keep the style of the enhanced images, which has also been pointed out in a recent work~\cite{chen2021scene}. Though HR images are not trustworthy, pixel information from HR images can help the model to enhance the input images, rather than totally regenerate them, which is a much more challenging and uncontrollable task. In HiREN, we use mean-squared-error (MSE) to compute pixel loss to keep the style unchanged. Formally, we have
% Charbonnier Loss~\cite{charbonnier1994two}, instead following recent works that utilize mean-squared-error (MSE), to compute a pixel loss to constraint the style to be unchanged. Mathematically, we have
\begin{equation}
\label{eq:sty}
    \mathcal{L}_{sty} = ||I_{HQ},I_{HR}||_2.
\end{equation}
% \begin{equation}
% \label{eq:sty}
%     \mathcal{L}_{sty} = \sqrt{(I_{HQ}-I_{HR})^2+\epsilon}.
% \end{equation}
% Comparing with MSE, Charbonnier Loss allow minor differences in detail by introducing the hyper-parameter $\epsilon$, while such differences can strengthen the recognizable ability of the enhancement image with the help of recognition loss.
With the recognition loss Eq.~(\ref{eq:rec}) and the style loss Eq.~(\ref{eq:sty}), the whole loss function of the HR enhancement branch can be written as follows:
\begin{equation}
\label{eq:hr}
    \mathcal{L}_{HR} = \alpha \mathcal{L}_{rec} + \mathcal{L}_{sty},
\end{equation}
where $\alpha$ is a hyper-parameter to trade-off the two losses.
% \subsection{Overall Loss Function}
%{\color{red}
\subsection{Quality Estimation Module}
%Although comparing with existing STISR techniques that directly take $I_{HR}$ to supervise the LR branch,
Though we can improve the quality of supervision information with the help of the HR enhancement branch, we cannot guarantee the correctness of the supervision information. Therefore, to suppress wrong supervision information, we design a quality estimation module $f_{QE}$ to evaluate the qualities of HQ images and weight the losses of HQ images according to their qualities.

Let the original loss of the LR branch be $\mathcal{L}_o \in \mathbb{R}^{B}$, where $B$ denotes the batch size. We adopt the Levenshtein similarity~\cite{levenshtein1966binary} between the $i$-th HQ image's recognition result $pred_i$ of a recognizer $R$ and the corresponding ground truth $gt_i$ to measure its quality, and then utilize the quality values of all HQ images to compute the final loss:
\begin{equation}
\label{eq:lr}
    \mathcal{L}_{LR} = \mathcal{L}_{o} [NS(pred_1, gt_1), ..., NS(pred_B, gt_B)]^\top / B,
\end{equation}
where $NS(\cdot, \cdot)$ denotes the Levenshtein similarity, which has the following two advantages: 1) its value falls between 0 and 1; 2) it has a smooth response, thus can gracefully capture character-level errors~\cite{biten2019scene}. These advantages make it suitable to weight the losses of HQ images.
%}

\begin{algorithm}[t]
\caption{The online usage of HiREN.}
\begin{algorithmic}[1]
\State {\bf Input:}
Training dataset $\mathcal{D}$ and a pretrained recognizer $R$
\State Initialize $f_{HR}$ and $f_{LR}$
\State \# Develop the HR enhancement branch
\While{$f_{HR}$ is not converged}
    \State $I_{HR},p_{GT}\sim \mathcal{D}$
    \State $I_{HQ}=f_{HR}(I_{HR})$
    \State Compute $\mathcal{L}_{rec}$ via Eq.~(\ref{eq:rec})
    \State Compute $\mathcal{L}_{sty}$ via Eq.~(\ref{eq:sty})
    \State $\mathcal{L}_{HR} = \alpha \mathcal{L}_{rec}+\mathcal{L}_{sty}$
    \State Optimize $f_{HR}$ via $\mathcal{L}_{HR}$
\EndWhile
\While{$f_{LR}$ is not converged}
    \State $I_{LR},I_{HR}\sim \mathcal{D}$
    \State $I_{HQ}=f_{HR}(I_{HR})$
    \State $I_{SR}=f_{LR}(I_{LR})$
    \State Compute $\mathcal{L}_{o}$ according to $I_{SR}$ and $I_{HQ}$
    \State Optimize $f_{LR}$ with respect to $\mathcal{L}_{o}$
\EndWhile
\State \Return $f_{LR}$
\end{algorithmic}
\label{alg:hire_online}
\end{algorithm}

\begin{algorithm}[t]
\caption{The offline usage of HiREN.}
\begin{algorithmic}[1]
\State {\bf Input:}
Training dataset $\mathcal{D}$ and the developed HR enhancement branch $f_{HR}$
\State Initialize $f_{LR}$
\State $\hat{\mathcal{D}} = \emptyset$
\For{$I_{LR},I_{HR}\sim \mathcal{D}$}
    \State $I_{HQ}=f_{HR}(I_{HR})$
    \State Add $(I_{HQ},I_{LR})$ to $\hat{\mathcal{D}}$
\EndFor
\While{$f_{LR}$ is not converged}
    \State $I_{HQ},I_{LR}\sim \hat{\mathcal{D}}$
    \State $I_{SR}=f_{LR}(I_{LR})$
    \State Compute $\mathcal{L}_{o}$ according to $I_{SR}$ and $I_{HQ}$
    \State Optimize $f_{LR}$ with respect to $\mathcal{L}_{o}$
\EndWhile
\State \Return $f_{LR}$
\end{algorithmic}
\label{alg:hire_offline}
\end{algorithm}
\subsection{The Usage of HiREN}
In this section, we introduce the usage of HiREN. As mentioned above, there are two ways to deploy it. One way is called ``online'', which can be easily implemented by plugged the HR enhancement branch to the training procedure of the LR recovery branch. The online installation algorithm of HiREN is given in Alg.~\ref{alg:hire_online}. As shown in Alg.~\ref{alg:hire_online}, the first thing we should do is to develop the HR enhancement branch (\textit{i.e.,} L4$\sim$L10). Specifically, given a STISR dataset $\mathcal{D}$, we first sample HR images and their corresponding text-level annotations from $\mathcal{D}$ (L5), then generate the enhanced images $I_{HQ}$ (L6). Finally, recognition loss and style loss described in Sec.~\ref{sec:loss:hr_branch} are computed to optimize the loss $f_{HR}$. After that, we plug the developed HR enhancement branch to the training procedure of the LR recover branch (L11$\sim$L16). In particular, after sampling LR and HR images from the dataset $\mathcal{D}$ (L12), we use the HR enhancement branch to generate the HQ image $I_{HQ}$ (L13). Finally, the HQ image, rather than the HR image used in typical works, and the SR image are utilized to compute the text-specific loss $\mathcal{L_{o}}$ to supervise the LR recovery branch (L11$\sim$L12).

The other way is called ``offline'', which can be implemented by caching all the enhanced HQ images. As can be checked in Alg.~\ref{alg:hire_offline}, after developing the HR enhancement branch $f_{HR}$, we sample all the LR-HR image pairs in the old dataset $\mathcal{D}$. Then, the corresponding HQ images are generated and then add to the new dataset $\hat{\mathcal{D}}$ (L6). During training the LR recovery branch, what we need to do is to sample LR-HQ image pairs to compute the loss $L_{o}$ for the optimization of the model. Such an installation does not introduce any additional training cost to the LR recovery branch. It is worth mentioning that the HR enhancement branch is removed during inference. That is, HiREN does not introduce any additional inference cost.

\section{Performance Evaluation}\label{sec:performance-evaluation}
In this section, we first introduce the dataset and metrics used in the experiments and the implementation details. Then, we evaluate HiREN and compare it with several state-of-the-art techniques to show its effectiveness and superiority. Finally, we conduct extensive ablation studies to validate the design of our method.

\subsection{Dataset and Metrics}
Two groups of datasets are evaluated in this paper: low-resolution scene text dataset TextZoom and regular scene text recognition datasets.
\subsubsection{Low-resolution scene text dataset}
The \textbf{TextZoom}~\cite{wang2020scene} dataset consists of 21,740 LR-HR text image pairs collected by lens zooming of the camera in real-world scenarios. The training set has 17,367 pairs, while the test set is divided into three settings based on the camera focal length: easy (1,619 samples), medium (1,411 samples), and hard (1,343 samples).

\subsubsection{Regular STR datasets}
These datasets are used to check the generalization power of our model trained on TextZoom when being adapted to other datasets. In particular, three regular STR datasets are evaluated in our paper to further check the advantage of HiREN: IC15-352~\cite{chen2021scene}, SVT~\cite{wang2011end}, and SVTP~\cite{phan2013recognizing}. In what follows, we give brief introductions on these datasets.

The \textbf{IC15-352} dataset is first divided in \cite{chen2021scene}. This dataset consists of 352 low-resolution images collected from the IC15~\cite{karatzas2015icdar} dataset.

Street View Text (\textbf{SVT})~\cite{wang2011end} is collected from the Google Street View. The test set contains 647 images. Many images in SVT are severely suffered from noise, blur, and low-resolution.

SVT-Perspective (\textbf{SVTP})~\cite{phan2013recognizing} is proposed for evaluating
the performance of reading perspective texts. Images in SVTP are picked from the side-view images in Google Street View. Many of them are heavily distorted by the non-frontal view angle. This dataset contains 639 images for evaluation.

The major metric used in this paper is word-level recognition accuracy that evaluates the recognition performance of STISR methods. Following the settings of previous
works~\cite{chen2022text}, we remove punctuation and convert uppercase letters to lowercase letters for calculating recognition accuracy. Besides, \textit{\textbf{Fl}oating-point \textbf{O}perations \textbf{P}er \textbf{S}econd} (FLOPS) is used to evaluate the computational cost of various methods. Following \cite{chen2022text,qin2022scene}, we only report \emph{Peak Signal-to-Noise Ratio} (PSNR) and \emph{Structure Similarity Index Measure} (SSIM)~\cite{wang2004image} as the auxiliary metrics to evaluate the fidelity performance because of the quality issue of the HR images.

\subsection{Implementation Details}
All experiments are conducted on 2 NVIDIA Tesla V100 GPUs with 32GB memory. The PyTorch version is 1.8. The HR enhancement branch is trained using Adam~\cite{kingma2014adam} optimizer with a learning rate of 0.0001. The batch
size $B$ is set to 48. The LR recovery branch is trained with the same optimizer and batch size but a higher learning rate of 0.001, which is suggested in \cite{ma2023text}. The recognizer $R$ used in our method is proposed in \cite{chen2021scene}. The hyper-parameters in HiREN are set as follows: $\alpha$ is set to 0.1, which is determined through grid search. The number of SRB blocks is set to 5 (\textit{i.e.,} $N=5$) and $C'$ is set to 32, which is the same as in \cite{wang2020scene}. The size of kernel $k$ is set to 32 (\textit{i.e.,} $d=32$), which is similar to that suggested in \cite{gu2019blind}. Our
training and evaluation are based on the following protocol: save the averagely best model during training with CRNN as the recognizer, and use this model to evaluate the other recognizers (MORAN, ASTER) and the three settings (easy, medium, hard).

\subsection{Performance Improvement on SOTA Approaches}
\begin{table*}[t]
%\begin{subtable}
\centering
\caption{Performance (recognition accuracy) improvement on TextZoom.}
\scalebox{1.0}{
\begin{tabular}{c||c c c c|c c c c|c c c c}
\hline
\multirow{2}*{Method}  &  \multicolumn{4}{c|}{CRNN \cite{shi2016end}} & \multicolumn{4}{c|}{MORAN \cite{luo2019moran}} & \multicolumn{4}{c}{ASTER \cite{shi2018aster}}\\
% \cline{3-14}
\cline{2-13}
~  & Easy & Medium & Hard & Average & Easy & Medium & Hard & Average & Easy & Medium & Hard & Average  \\
\hline
\hline
LR & 37.5\% & 21.4\% & 21.1\% & 27.3\% & 56.2\% & 35.9\% & 28.2\% & 41.1\% & 64.6\% & 42.0\% & 31.7\% & 47.2\%  \\
+HiREN & \textbf{37.7}\% & \textbf{27.9}\% & \textbf{23.5}\% & \textbf{30.2}\% & \textbf{57.9}\% & \textbf{38.2}\% & \textbf{28.7}\% & \textbf{42.6}\%  &\textbf{66.4}\% &\textbf{43.4}\% &\textbf{32.3}\% &\textbf{48.5}\%  \\
\hline
HR  & 76.4\% & 75.1\% & 64.6\% & 72.4\% & \textbf{89.0}\% & 83.1\% & 71.1\% & 81.6\% & 93.4\% & 87.0\% & 75.7\% & 85.9\%  \\
+HiREN  & \textbf{77.5}\% & \textbf{75.4}\% & \textbf{65.0}\% & \textbf{72.9}\% & 88.8\% & \textbf{83.7}\% & \textbf{71.9}\% & \textbf{82.0}\% & \textbf{93.5}\% & \textbf{87.5}\% & \textbf{76.2}\% & \textbf{86.3}\% \\
\hline
\hline
SRCNN & 39.8\% & 23.4\% & 21.7\% & 29.0\% & 57.7\% & 36.1\% & 28.5\% & 41.8\% & 65.5\% & 41.9\% & 31.7\% & 47.5\%    \\
+HiREN & \textbf{41.6}\% & \textbf{24.0}\% & \textbf{23.7}\% & \textbf{30.4}\% & \textbf{61.1}\% & \textbf{38.6}\% & \textbf{29.3}\% & \textbf{44.0}\% & \textbf{67.5}\% & \textbf{44.7}\% & \textbf{32.8}\% & \textbf{49.5}\%    \\
\hline
TSRN & 52.8\% & 39.8\% & 31.6\% & 42.1\% & 64.5\% & 49.3\% & 36.7\% & 51.1\% & 69.7\% & 54.8\% & 41.3\% & 56.2\% \\
+HiREN & \textbf{56.5}\% & \textbf{44.1}\% & \textbf{32.2}\% & \textbf{45.0}\% & \textbf{68.5}\% & \textbf{52.5}\% & \textbf{38.6}\% & \textbf{54.2}\% & \textbf{73.5}\% & \textbf{56.3}\% & \textbf{39.2}\% & \textbf{57.4}\%    \\
% \hline
% \cite{nakaune2021skeleton} & BMVC21 & 56.5\%  & 42.3\%  & 32.5\%  & 44.5\%  & 72.9\%  & 55.9\%  & 40.6\%  & 57.5\%  & 77.3\%  & 59.6\%  & 42.7\%  & 60.9\% \\
% STT & 59.2\% & 44.2\% & 36.0\% & 47.3\% & 70.3\% & 54.5\% & 41.2\% & 56.3\% & 75.1\% & 59.3\% & 42.1\% & 59.9\%  \\
% +HiREN & \textbf{61.6}\% & \textbf{49.8}\% & \textbf{37.2}\% & \textbf{50.3}\% & \textbf{73.4}\% &\textbf{56.8}\% & \textbf{41.6}\% & \textbf{58.3}\% & \textbf{77.5}\% & \textbf{60.7}\% & \textbf{44.5}\% & \textbf{62.0}\%    \\
% \hline
% PCAN~\cite{zhao2021scene} & 59.6\% & 45.4\% & 34.8\% & 47.4\% & 73.7\% & 57.6\% & 41.0\% & 58.5\% & 77.5\% & 60.7\% & 43.1\% & 61.5\%  \\
% +HiREN & 46.8\% & 27.9\% & 26.5\% & 34.5\% & 63.1\% & 42.9\% & 33.6\% & 47.5\% & 67.3\% & 46.6\% & 35.1\% & 50.7\%    \\
% \hline
% \hline
% TPGSR& -\% & -\% & -\% & -\% & -\% & -\% & -\% & -\% & -\% & -\% & -\% & -\%  \\
% +HiREN & \textbf{63.7}\% & 49.5\% & 35.7\% & \textbf{50.5}\% & 70.9\% & 55.9\% & 39.5\% & 56.4\%  & 76.2\% & 60.0\% & 41.8\% & 60.4\%  \\
\hline
TG  & 60.5\% & 49.0\% & 37.1\% & 49.6\% & 72.0\% & 57.6\% & 40.0\% & 57.6\% & 76.0\% & 61.4\% & 42.9\% & 61.2\%  \\
+HiREN & \textbf{62.4}\% & \textbf{51.2}\% & \textbf{37.5}\% & \textbf{51.1}\% & \textbf{73.4}\% & \textbf{58.4}\% & \textbf{41.0}\% & \textbf{58.6}\% & \textbf{77.5}\% & \textbf{61.5}\% & \textbf{43.0}\% & \textbf{61.7}\%    \\
\hline
TPGSR & 63.1\% &52.0\% & 38.6\% & 51.8\% & \textbf{74.9}\% & 60.5\% & 44.1\% & \textbf{60.5}\%  & \textbf{78.9}\% & 62.7\% & 44.5\% & 62.8\%  \\
+HiREN & \textbf{63.5}\% & \textbf{52.7}\% & \textbf{38.8}\% & \textbf{52.4}\% & 74.7\% & \textbf{60.9}\% & \textbf{44.1}\% & \textbf{60.5}\%  & 78.3\% & \textbf{63.5}\% & \textbf{45.6}\% & \textbf{63.5}\%  \\
\hline
\end{tabular}}
\label{tab:final result}
\end{table*}
\label{sec:exp:sota}

\subsubsection{Recognition performance improvement}
Here, we evaluate our method on \textbf{TextZoom}. Since HiREN is a framework that can work with most existing methods, we plug HiREN to the training of several typical super-resolution methods to check the universality and effectiveness of HiREN, including one generic method SRCNN~\cite{dong2015image}, two recent proposed STISR methods TSRN~\cite{wang2020scene}, TG~\cite{chen2022text}, and one iterative-based and clue-guided STISR method TPGSR~\cite{ma2023text}. To show that HiREN can support various recognizers, we follow previous works~\cite{ma2023text,chen2021scene,chen2022text} and evaluate the recognition accuracy on three recognizers: CRNN~\cite{shi2016end}, MORAN~\cite{luo2019moran} and ASTER~\cite{shi2018aster}. We re-implement these methods to unify hardware, software, and evaluation protocols for fair comparison. Generally, our results are higher than those in the original papers. For example, with CRNN the averaged accuracy of TG is boosted from 48.9\% to 49.6\%. All the results are presented in Tab.~\ref{tab:final result}.

We first check the universality of HiREN. As can be seen in Tab.~\ref{tab:final result}, HiREN significantly boosts the recognition performance in almost all the cases, except for one case on TPGSR, which means that HiREN can work well with various existing techniques. As for the performance improvement of HiREN, taking a non-iterative method for example. The state-of-the-art TG~\cite{chen2022text} achieves 49.6\%, 57.6\% and 61.2\% averaged accuracy respectively with the three recognizers (see the 9th row). After equipping our method HiREN, the accuracy is lifted to 51.1\%, 58.6\% and 61.7\% (increasing by 1.5\%, 1.0\%, and 0.5\%) respectively (see the 10th row). This demonstrates the effectiveness of our method. Results on more datasets and recognizers are given in the supplementary materials to demonstrate its universality.

It is worth mentioning that our HR enhancement branch can also be applied to weakly supervising the enhancement of LR and HR images to lift their recognition accuracies, as shown in the 3rd and 5th rows of Tab.~\ref{tab:final result}. This further supports the universality of our technique. Results above show the promising application potential of our method --- not only work with STISR methods, but also pioneer weakly supervised enhancement of LR and HR text images.

\begin{table}[t]
%\begin{subtable}
\centering
\caption{Performance comparison on three STR datasets with CRNN as recognizer.}
\scalebox{1.0}{
\begin{tabular}{c|ccc}
\hline
Method & IC15-352 & SVT & SVTP  \\
\hline
% LR & 81.0\% & 79.1\% & 63.3\% &-  \\
LR & 49.4\% & 74.8\% & 60.8\%  \\
\hline
TSRN & 48.9\% & 72.6\% & \textbf{61.4}\% \\
+HiREN & \textbf{52.3}\% & \textbf{74.8}\% & 60.3\% \\ %rec only
\hline
TG & 59.1\% & 74.2\% & 60.2\% \\
+HiREN & \textbf{61.7}\% & \textbf{76.5}\% & \textbf{60.5}\% \\ %rec only
\hline
TPGSR & 66.2\% & 77.4\% & 62.8\% \\ % ling only
+HiREN & \textbf{66.8}\% & \textbf{78.7}\% & \textbf{63.6}\%  \\ % vis only
\hline
\end{tabular}}
\label{tab:normal}
\end{table}

Furthermore, to better demonstrate the universality of HiREN, we conduct more experiments on more STR datasets and recently proposed STR datasets. We first evaluate our method on three STR datasets, including IC15-352, SVT, and SVTP. We use the STISR models (TSRN, TG, TPGSR, and our techniques performed on them) developed on the TextZoom dataset to evaluate these datasets. The experimental results on IC15-352, SvT, and SVTP are given in Tab.~\ref{tab:normal}. As shown in Tab.~\ref{tab:normal}, HiREN also works well on them and achieve lifted performance in almost all the cases. In particular, the performance of TPGSR on three datasets are lifted from 66.2\%, 77.4\%, 62.8\% to 66.8\%, 78.7\%, and 63.6\%, respectively, which demonstrates the advantage of HiREN.
\begin{table}[t]
%\begin{subtable}
\centering
\caption{Performance of recent recognizers on TextZoom.}
\scalebox{1.0}{
\begin{tabular}{c|cc}
\hline
Method & SEED~\cite{qiao2020seed} & ABINet~\cite{fang2021read} \\
 \hline
LR & 45.8\% & 61.0\% \\
HR & 84.8\% & 89.8\% \\
 \hline
TSRN & 56.3\% & \textbf{64.0}\%   \\
+HiREN & \textbf{56.5}\% & 63.8\% \\ %rec only
 \hline
TG & 60.7\% & \textbf{66.0}\%   \\
+HiREN & \textbf{60.9}\% & 65.9\% \\ %rec only
\hline
TPGSR & 61.7\% & 67.5\% \\ % ling only
+HiREN & \textbf{62.2}\% & \textbf{68.1}\% \\ % vis only
\hline
\end{tabular}}
\label{tab:rec}
\end{table}

Apart from that, we also give the experimental results on more recently proposed recognizers including SEED~\cite{qiao2020seed} and ABINet~\cite{fang2021read}. The experimental results are given in Tab.~\ref{tab:rec}.
As can be checked in Tab.~\ref{tab:rec}, these recent recognizers still find difficulty in recognizing low-resolution text images. For example, SEED and ABINet can only correctly read 45.8\% and 61.0\% of LR images, which are inferior to performance of reading HR images (\textit{i.e.,} 84.8\% and 89.8\%). Our method HiREN can also achieve boosted performance on these recognizers in almost all the cases.
		
\begin{table}[t]
\centering
\caption{Fidelity and recognition results on major existing methods. The results are obtained by averaging three settings (easy, medium and hard).}
\scalebox{0.95}{
	\begin{tabular}{c||cc|cc|c}
		\hline
		\multirow{3}*{Method} & \multicolumn{5}{c}{Metrics}\\
		\cline{2-6}
		~&\multicolumn{2}{c|}{SR-HR}&\multicolumn{2}{c|}{SR-HQ}& Avg \\
		\cline{2-5}
		~& PSNR & SSIM($\times 10^{-2}$)& PSNR & SSIM($\times 10^{-2}$) & Acc  \\
		\hline
		\hline
		LR & 20.35 & 69.61 & 20.73 &68.76 &27.3\%\\
		\hline
		TSRN &21.84 &76.34& 21.08 & 74.76& 42.1\%\\
		+HiREN &\textbf{22.01} &\textbf{76.60}& \textbf{21.46} & \textbf{76.23} & \textbf{45.0}\%\\
		\hline
		TG & \textbf{21.47} & \textbf{73.57} &\textbf{20.89} &72.59& 49.6\%\\
		+HiREN & 21.12 & 73.43 &20.84 &\textbf{73.78}& \textbf{51.1}\%\\
		\hline
		\hline
		TPGSR & \textbf{22.05} & \textbf{76.71}&21.05
 &\textbf{76.77} & 51.8\%\\
		+HiREN & 21.69 & 75.97&\textbf{21.15} &76.44 & \textbf{52.4}\%\\
		\hline
\end{tabular}}
\label{tab:psnr}
\end{table}
 \begin{figure*}[t]
	\begin{center}
		\includegraphics[width=1.0\linewidth]{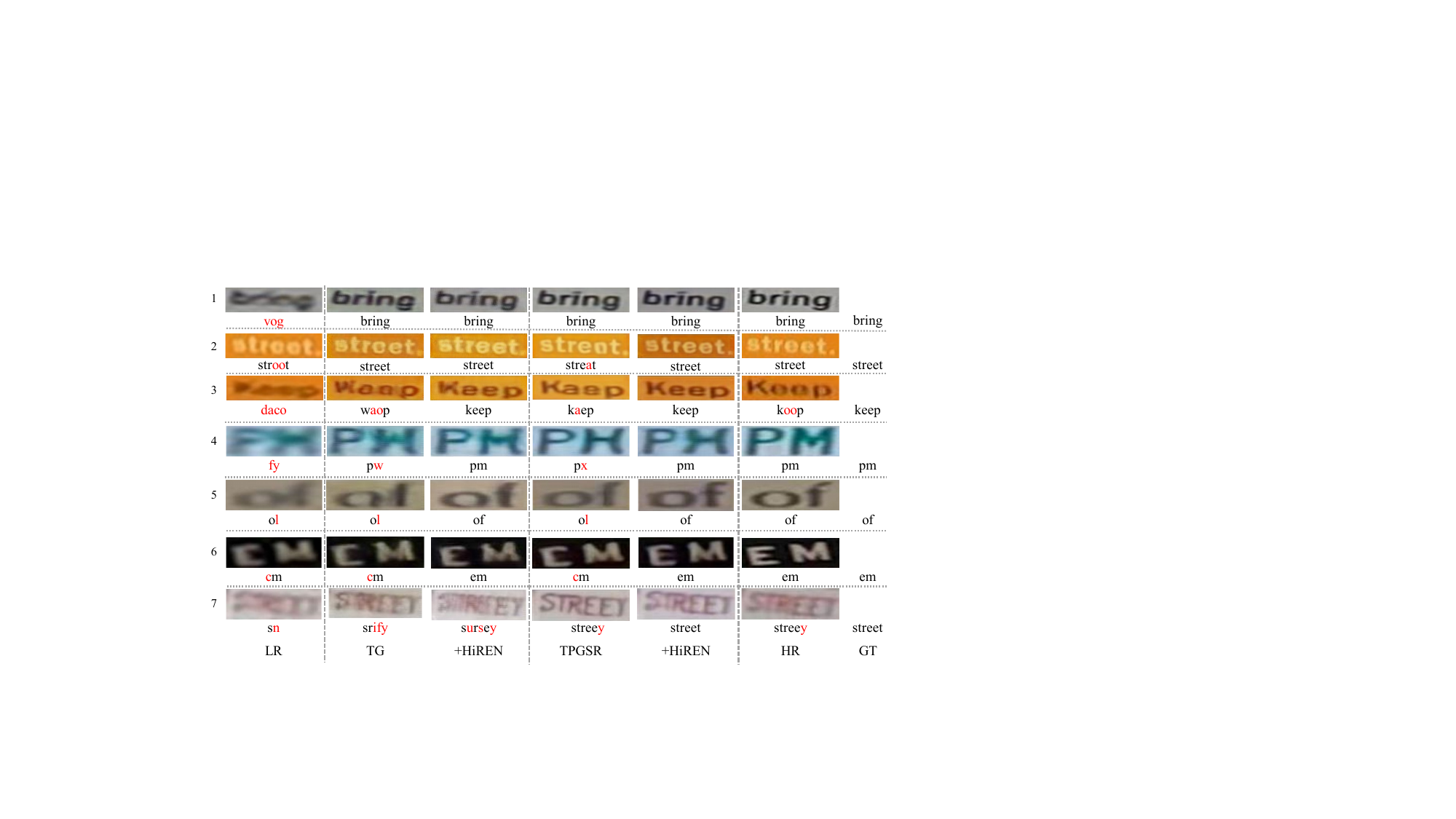}
	\end{center}
	\caption{Examples of generated images. Here, GT indicates ground truth. We use CRNN as the recognizer. Red/black characters indicate incorrectly/correctly recognized.}
	\label{fig-vis}
\end{figure*}
\subsubsection{Fidelity improvement}
We also report the results of fidelity improvement (PSNR and SSIM) on major existing methods in Tab.~\ref{tab:psnr}. Notice that these fidelity metrics have the following limitations. On the one hand, PSNR and SSIM globally measure the similarity between SR image and the ground truth image, including both characters and background. With the goal of lifting the recognition ability and readability of the scene text images, STISR should put more emphasis on recovering characters rather than the background~\cite{chen2022text,qin2022scene}. On the other hand, as pointed out by our paper, HR images are suffering various quality issues. Ergo, it is inappropriate to measure the pixel similarity between erroneous HR images whose pixels are not trustworthy. Therefore, we only present PSNR and SSIM as auxiliary metrics to roughly draw some conclusions.

Notice that existing methods utilize SR-HR image pairs to calculate PSNR and SSIM. However, as mentioned above, the HR images are suffering from quality issues. Hence, we additionally provide the fidelity results of calculating PSNR and SSIM between SR and HQ images. The experimental results are given in Tab.~\ref{tab:psnr}. As can be seen in Tab.~\ref{tab:psnr}, 1) A higher PSNR does not means a higher recognition accuracy. For example, the PSNR of TG in SR-HR is inferior to that of TSRN (\textit{i.e.,} 21.47 v.s. 21.84) but TG performs better on recognition accuracy (\textit{i.e.,} 49.6\% v.s. 42.1\%). The reason lies in that TG is a stroke-focused technique, focusing on recovering fine-grained stroke details rather than the whole image quality including background that is minor to recognition. This is consistent with the results in~\cite{chen2022text}. 2) Comparing with the original models, after applying HiREN, the SR-HQ fidelity performance of the new models are boosted in almost all cases. 3) HiREN gets a low performance on the PSNR and SSIM of SR-HR images but obtains an improved recognition performance, which supports the quality issue of HR images.

\subsubsection{Visualization}

Here, we visualize several examples in Fig.~\ref{fig-vis} to better demonstrate the performance of our technique. We can see that HiREN can help the existing methods to recover the blurry pixels better (see the 2nd $\sim$ 6th cases). In particular, a better ``ee'' in the 2nd and 3rd cases, `m' in the 4th case, `f' in the 5th case, and `e' in the 6th case are obtained by our technique. Besides, in some extremely tough cases where even with the HR images the recognition is hard, HiREN can still achieve better recovery (see the 7th case). These results show the power of HiREN.
% We also provide a bad case to illustrate the limitations of recent methods and HiREN. As can be seen in the 5th case, recent methods usually rely on a vocabulary~\cite{wan2020vocabulary}, which makes the models guess the blurry pixels via the corpus that can be learned from the training dataset. This degrades the models' ability to recover numbers and punctuation.

\begin{table}[t]
%\begin{subtable}
\centering
\caption{The training and inference costs of our method. The cost is measured by the FLOPs(G).}
\scalebox{1.0}{
\begin{tabular}{c|cc}
\hline
\multirow{2}*{Method} & \multicolumn{2}{c}{Metrics}\\
\cline{2-3}
~& Training cost & Inference cost\\
\hline
TG & 19.60 & 0.91\\
+HiREN(Online) & 20.59 & 0.91\\
+HiREN(Offline) & 19.60& 0.91\\
\hline
TPGSR & 7.20 &7.20\\
+HiREN(Online) &8.19&7.20\\
+HiREN(Offline) &7.20&7.20\\
 \hline

\end{tabular}}
\label{tab:speed}
\end{table}

\subsubsection{Training and inference cost}
%{\color{red}
We have discussed the high performance of our technique above. In this section, we provide the results of training and inference costs to show the efficiency of HiREN. Specifically, We take TG and TPGSR as baselines and add HiREN to them and count their FLOPS during training and inference. The experimental results are presented in Tab.~\ref{tab:speed}. In terms of training cost, we can see that the offline deployment of HiREN does not incur any additional cost. As for online version, we can see that the additional computational cost caused by HiREN is negligible (\textit{e.g,} from 19.60G to 20.59G, only 0.99G). What is more, neither of the two variants introduce any additional inference cost. In conclusion, the offline deployment not only saves training and inference cost, but also significantly boosts the performance. These results validate the efficiency of our method.

\subsection{Ablation Study}
We conduct extensive ablation studies to validate the
design of our method. Since our method is designed to enhance HR images during training, the metric used in this section is the recognition accuracy measured by the average accuracy of CRNN on training set, denoted as $Acc_{train}$.
\begin{table}[t]
%\begin{subtable}
\centering
\caption{The ablation studies of the HR enhancement branch. Here, \XSolidBrush means the corresponding module is not applied, and Charb is Charbonnier Loss~\cite{charbonnier1994two}.}
\scalebox{1.0}{
\begin{tabular}{c|c|cc|c}
\hline
\multirow{2}*{ID} & \multirow{2}*{Kernel-guided} &\multicolumn{2}{c|}{Loss functions} & \multirow{2}*{$Acc_{train}$}\\
\cline{3-4}
~&~& $\mathcal{L}_{rec}$ & $\mathcal{L}_{sty}$ & ~ \\
\hline
\hline
1& \XSolidBrush & \XSolidBrush & \XSolidBrush & 66.9  \\
\hline
2&\XSolidBrush & $\checkmark$ & MSE & 72.7  \\
3&$\checkmark$ & $\checkmark$ & \XSolidBrush & 66.1  \\
4&$\checkmark$ & \XSolidBrush & MSE & 67.4  \\
5&$\checkmark$ & $\checkmark$ & Charb & 67.5  \\
6&$\checkmark$ & $\checkmark$ & L1 & 67.3 \\
7&$\checkmark$ & $\checkmark$ & MSE & 74.1 \\
\hline
\end{tabular}
}
\label{tab:ab}
\end{table}

\subsubsection{Design of the HR enhancement branch}
Here, we check the design of the HR enhancement branch. As mentioned above, two techniques are developed to promote the enhancement of HR images: kernel-guided enhancement network $f_{ke}$ and the loss $\mathcal{L}_{HR}$. We conduct experiments to check their effects. The experimental results are presented in Tab.~\ref{tab:ab}. Visualization of the effect of the HR enhancement branch is given in the supplementary materials.

\emph{The effect of the HR enhancement branch}. Comparing the results in the 1st and 7th rows of Tab.~\ref{tab:ab}, we can see that the HR enhancement branch lifts the accuracy from 66.9\% to 74.1\%, which proves the effect of the branch as a whole.

\emph{The effect of kernel-guided enhancement network}. To check the power of the kernel-guided enhancement network, we design a variant that removes the kernel predictor. Comparing the results of the 2nd and 7th rows in Tab.~\ref{tab:ab}, we can see that the variant without the kernel predictor is inferior to that with the kernel predictor (72.7\% v.s. 74.1\%). This demonstrates the effectiveness of the proposed kernel-guided enhancement network.

\emph{The design of loss function}. Here, we check the design of the loss function used in the HR enhancement branch. We first remove the recognition loss $\mathcal{L}_{rec}$ and the style loss $\mathcal{L}_{sty}$ separately. As can be seen in the 3rd, 4th, and 7th rows in Tab.~\ref{tab:ab}, comparing with the combined one, the performance of using only one single loss is degraded. Next, we check the selection of style loss. Specifically, we consider three candidates (MSE, Charbonnier and L1) for the style loss function. As can be seen in the 5th, 6th, and 7th rows of Tab.~\ref{tab:ab}, MSE loss outperforms Charbonnier loss~\cite{charbonnier1994two} and L1 loss. The reason lies in that MSE penalizes large errors and is more tolerant to small errors, which is more suitable for HiREN to enhance the blurry or missed character details and keep the style unchanged~\cite{zhao2016loss}. Ergo, MSE is selected as the style loss in HiREN.

\begin{table}[t]
%\begin{subtable}
\centering
\caption{The determination of $\alpha$. The metric is $Acc_{train}$.}
\scalebox{1.0}{
\begin{tabular}{c|ccccccc}
\hline
\multirow{2}*{Metric} & \multicolumn{6}{c}{$\alpha$}\\
\cline{2-8}
~ & 0.5 & 0.2 & 0.1 & 0.05 & 0.025 &0.01 & 0.005  \\
% & 0.001 & 0.0005 & 0.0002  \\
 \hline
$Acc_{train}$ &73.6 & 73.4 &\textbf{74.1} & \textbf{74.1} &72.3& 72.2 &71.2 \\
% & 68.6 & 68.8 & 67.5 \\
\hline
\end{tabular}}
\label{tab:alpha}
\end{table}

\subsubsection{Hyper-parameter study}
Here, we provide the grid search results of the hyper-parameter $\alpha$ introduced in HiREN for balancing the two losses. The results are presented in Tab.~\ref{tab:alpha}. As can be seen in Tab.~\ref{tab:alpha}, the best performance is achieved when  $\alpha$=0.1 and 0.05.
\begin{table}[t]
%\begin{subtable}
\centering
\caption{Ablation study on the quality estimation module. The metric is the recognition accuracy of CRNN on the test set of TextZoom.}
\scalebox{1.0}{
\begin{tabular}{c|cccc}
\hline
Method & SRCNN &TSRN &TG & TPGSR\\
\cline{1-5}
% & 0.001 & 0.0005 & 0.0002  \\
 \hline
without $f_{QE}$ &30.2\% &44.2\% &51.0 &51.9\% \\
with $f_{QE}$& \textbf{30.4}\% &\textbf{45.0}\% &\textbf{51.1} &\textbf{52.4}\%\\
\hline
\end{tabular}}
\label{tab:qe}
\end{table}
%{\color{red}
\subsubsection{The effect of loss quality estimation module}
Here, we compare the performances of different models w/o the quality estimation module. As can be seen in Tab.~\ref{tab:qe}, without $f_{QE}$, all methods are degraded, which demonstrates the effect of the quality estimation module.%}

\section{Discussion}\label{sec:discussion}
In this section, we discuss some issues to better demonstrate the advantages of HiREN and point out some limitations of the proposed method.

\begin{figure}[t]
	\begin{center}
		\includegraphics[width=1.0\linewidth]{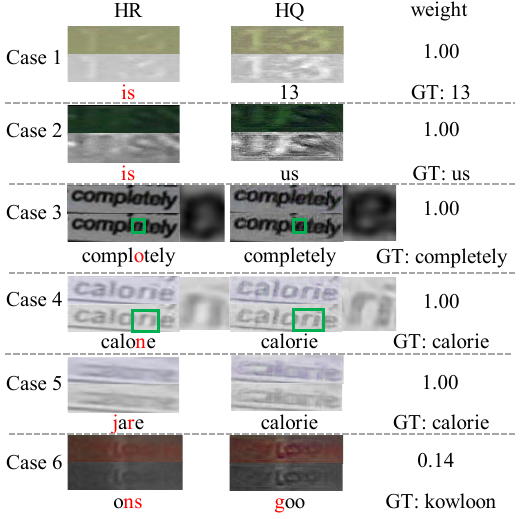}
	\end{center}
	\caption{Some examples of low-quality HR images, their enhanced results (HQ) by our method, and their weight calculated by the quality estimation module, as well as the recognized results. For each case, the 1st row shows HR and HQ images, the 2nd row presents the normalized HR and HQ images to highlight their visual differences, and the 3rd row gives the recognized characters: red indicates incorrectly recognized, and black means correctly recognized.}
	\label{fig-error}
\end{figure}
\subsection{Which kind of quality issues do HR images have?}
We conduct a visualization study to demonstrate the quality issues of HR images. As can be checked in Fig.~\ref{fig-error}, HR images are suffering from including but not limited to low-contrast (1st, 2nd and 6th cases), blurry (3rd and 4th cases) and motion blur (5th case). These unknown degradations obviously threaten the recognition of HR images and subsequently provide erroneous supervision to the recovery of the LR images.

\subsection{How does HiREN lift the quality of supervision information?}
To cope with various quality problems of HR images, HiREN generates HQ images through different strategies. In particular, HiREN makes the texts more prominent to solve low-contrast (e.g. the 1st and 2nd cases in Fig.~\ref{fig-error}). With respect to the blurry issue, HiREN makes the incorrectly recognized texts more distinguishable (e.g. ``e'' in the 3rd case and ``ri'' in the 4th case in Fig.~\ref{fig-error}). HiREN also tries to reduce the motion blur in the 5th case of Fig.~\ref{fig-error}. Although in some tough cases, HiREN fails to generate a correct HQ image (e.g. the 6th case in Fig.~\ref{fig-error}), our quality estimation module weights its loss to a small value to suppress the erroneous supervision information.

\begin{figure}[t]
	\begin{center}
		\includegraphics[width=1.0\linewidth]{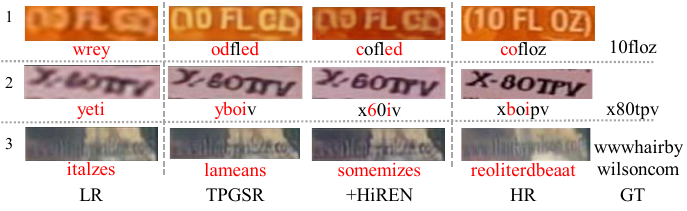}
	\end{center}
	\caption{Error analysis of HiREN. Here, GT indicates ground truth. We use CRNN as the recognizer. Red/black characters indicate incorrectly/correctly recognized.}
	\label{fig-error-ana}
\end{figure}

\subsection{Error Analysis}
\label{sec:error-analy}
In this section, we perform an error analysis of HiREN to provide possible research directions for further works.
Concretely, we provide some error cases in Fig.~\ref{fig-error-ana} to illustrate the limitations of recent works and HiREN. As can be seen in the 1st$\sim$2nd cases, recent methods usually rely on a vocabulary~\cite{wan2020vocabulary}, which makes the models guess the blurry pixels via the corpus that can be learned from the training dataset. This degrades the models' ability to recover numbers and punctuation. As a result, although HiREN recovers more characters than the original TPGSR, the word-level recovery still fails. Besides, as shown in the 3rd case, in some tough cases where the LR and HR images are extremely difficult to read, TPGSR and HiREN also fail to effectively do the recovery. This indicates the challenge of STISR.
% \subsection{The mertics of PSNR and SSIM.}
% As mentioned above, recent STISR papers have pointed out the limitation of two fidelity metrics: PSNR and SSIM. A higher fidelity performance means the generated SR image is more similar to HR images.
% \section{Discussions}
% \subsection{How Can We Further Improve HiREN?}
% In this paper, HiREN is designed in a simple yet efficient manner. That is, we train one model to conduct enhancement for all HR images. This ensures the efficiency of our technique, as shown in Tab.~\ref{tab:speed}. Here, we present some directions to further boost the enhancement ability of HiREN. Concretely, 1) Replace the recognizer $R$ with a more powerful one for more accurate supervision; 2) Search tailored parameters for each HR image; 3) Directly feed the text-level annotation to the HR enhancement branch. Nevertheless, these strategies also have some shortcomings. For 1) and 2), the efficiency of HiREN will be significantly hurt. As for 3), this solution may make HiREN unable to enhance text images that lack text-level annotations, which restricts the application of HiREN. By jointly taking efficiency, performance, and application into consideration, we should consider more powerful alternatives.

\subsection{Limitations of HiREN}
On the one hand, HiREN may introduce some noise to the HR images and worsen their quality. However, such noise is very minor compared to the advantage brought by HiREN. Specifically, we find that 9,565 erroneously recognized images in TextZoom dataset are successfully enhanced by HiREN, which leads to correct recognition results, while only 128 images are deteriorated from correct to wrong. On the other hand, the training of the HR enhancement branch requires the feedback of a scene text recognizer and text-level annotations. This indicates that HiREN still needs some weak supervision information for supervision.

%\subsection{Limitations and Future Work}

\section{Conclusion}\label{sec:conclusion}
In this paper, we present a novel framework called HiREN to boost STISR performance. Different from existing works, HiREN aims at generating high-quality text images based on high-resolution images to provide more accurate supervision information for STISR. Concretely, recognizing the difficulty in catching the degradation from HQ to HR and obtaining the supervision information from HR images, we explore degradation kernel-guided super-resolution and the feedback of a recognizer as well as text-level annotations as weak supervision to train a HR enhancement branch. What is more, to suppress erroneous supervision information, a novel quality estimation module is designed to evaluate the qualities of images, which are used to weight their losses.
%Comparing with the existing techniques, in addition to higher performance, HiREN has three advantages: general, easy-to-use, and efficient.
Extensive experiments demonstrate the universality, high-performance and efficiency of HiREN. Our work provides a new solution for the STISR task. 

In the future, we will try to explore more advanced models to further advance the proposed technique. One the one hand, we will try to further improve the recovery ability of the HR enhancement branch or address the vocabulary reliance issue. On the other hand, we plan to apply HiREN to self-supervised or unsupervised settings when the recognizer and text-level annotations are not trustworthy or text-level annotations are lack during training. Last but not least, we will extend the idea of the proposed quality enhancement branch to build a new noisy learning algorithm for STISR.

% \section*{Acknowledgment}


\ifCLASSOPTIONcaptionsoff
  \newpage
\fi



\bibliographystyle{IEEEtran}
\bibliography{hire}
\end{document}